\documentclass[letterpaper, 10 pt, conference]{ieeeconf}
\IEEEoverridecommandlockouts
\usepackage{graphics} 
\usepackage{amsmath} 
\usepackage{amssymb}  
\usepackage{physics}
\usepackage{cite}
\usepackage{amsfonts}
\usepackage{esdiff}
\usepackage{xcolor}
\usepackage{siunitx}
\usepackage{graphicx}
\usepackage[font={footnotesize}]{caption}
\usepackage{float}
\usepackage{subfig}
\usepackage{flushend}
\usepackage{algorithm}
\usepackage[noend]{algpseudocode}

\usepackage{amsthm}
\theoremstyle{definition}
\graphicspath{ {./img/} }
\def\BibTeX{{\rm B\kern-.05em{\sc i\kern-.025em b}\kern-.08em
    T\kern-.1667em\lower.7ex\hbox{E}\kern-.125emX}}    

\title{\LARGE \bf Constrained Bandwidth Observation Sharing for Multi-Robot Navigation in Dynamic Environments via Intelligent Knapsack}

\author{Anirudh Chari$^{1}$, Rui Chen$^{2}$, Han Zheng$^{1}$, and Changliu Liu$^{2}$
\thanks{$^{1}$A. Chari and H. Zheng are with the Massachusetts Institute of Technology in Cambridge, MA, USA. (\tt{\{anichari,hanzheng\}@mit.edu})}%
\thanks{$^{2}$R. Chen and C. Liu are with the Robotics Institute at Carnegie Mellon University in Pittsburgh, PA, USA. (\tt{\{ruic3,cliu6\}@andrew.cmu.edu})}%
\thanks{This work is partially supported by the National Science Foundation, Grant No. 2144489.}%
}

\begin{document}

\maketitle
\thispagestyle{empty}
\pagestyle{empty}

\begin{abstract}
Multi-robot navigation is increasingly crucial in various domains, including disaster response, autonomous vehicles, and warehouse and manufacturing automation.
Robot teams often must operate in highly dynamic environments and under strict bandwidth constraints imposed by communication infrastructure, rendering effective observation sharing within the system a challenging problem.
This paper presents a novel optimal communication scheme, Intelligent Knapsack (iKnap), for multi-robot navigation in dynamic environments under bandwidth constraints.
We model multi-robot communication as belief propagation in a graph of inferential agents.
We then formulate the combinatorial optimization for observation sharing as a 0/1 knapsack problem, where each potential pairwise communication between robots is assigned a decision-making utility to be weighed against its bandwidth cost, and the system has some cumulative bandwidth limit. 
We evaluate our approach in a simulated robotic warehouse with human workers using ROS2 and the Open Robotics Middleware Framework.
Compared to state-of-the-art broadcast-based optimal communication schemes, iKnap yields significant improvements in navigation performance with respect to scenario complexity while maintaining a similar runtime. 
Furthermore, iKnap utilizes allocated bandwidth and observational resources more efficiently than existing approaches, especially in very low-resource and high-uncertainty settings.
Based on these results, we claim that the proposed method enables more robust collaboration for multi-robot teams in real-world navigation problems.

\end{abstract}


\section{Introduction}\label{intro}
The rapid proliferation of multi-robot systems across various domains has underscored the critical importance of efficient communication strategies to navigate the environment and coordinate actions \cite{gielis}. 
Autonomous vehicles can leverage over-the-air communication to share information about the road environment, yielding safer transportation systems \cite{ye,liu}.
Search-and-rescue robot teams deployed in hazardous areas must efficiently relay real-time observations to locate survivors and assess damage \cite{queralta}.
In industrial environments such as manufacturing and warehouses, robot groups coordinate complex tasks such as material handling and inventory management \cite{dacostabarros,schuster}.

Much of the current work in multi-robot coordination assumes perfect communication within the system.
However, in all the aforementioned settings, bandwidth limitations imposed by centralized or decentralized communication infrastructure are inherent and inevitable \cite{jawhar}.
Moreover, the increasingly complex and dynamic nature of the tasks and environments in which multi-robot systems are deployed will demand greater levels of coordination, and thus more communication \cite{gielis}.
This necessitates the treatment of bandwidth as a luxury rather than a guarantee to ensure the feasibility of real-world implementation of these systems.
Thus, we are motivated to pursue \textit{utility-conscious communication schemes} for robot teams.

\begin{figure}[t]
\includegraphics[width=\linewidth]{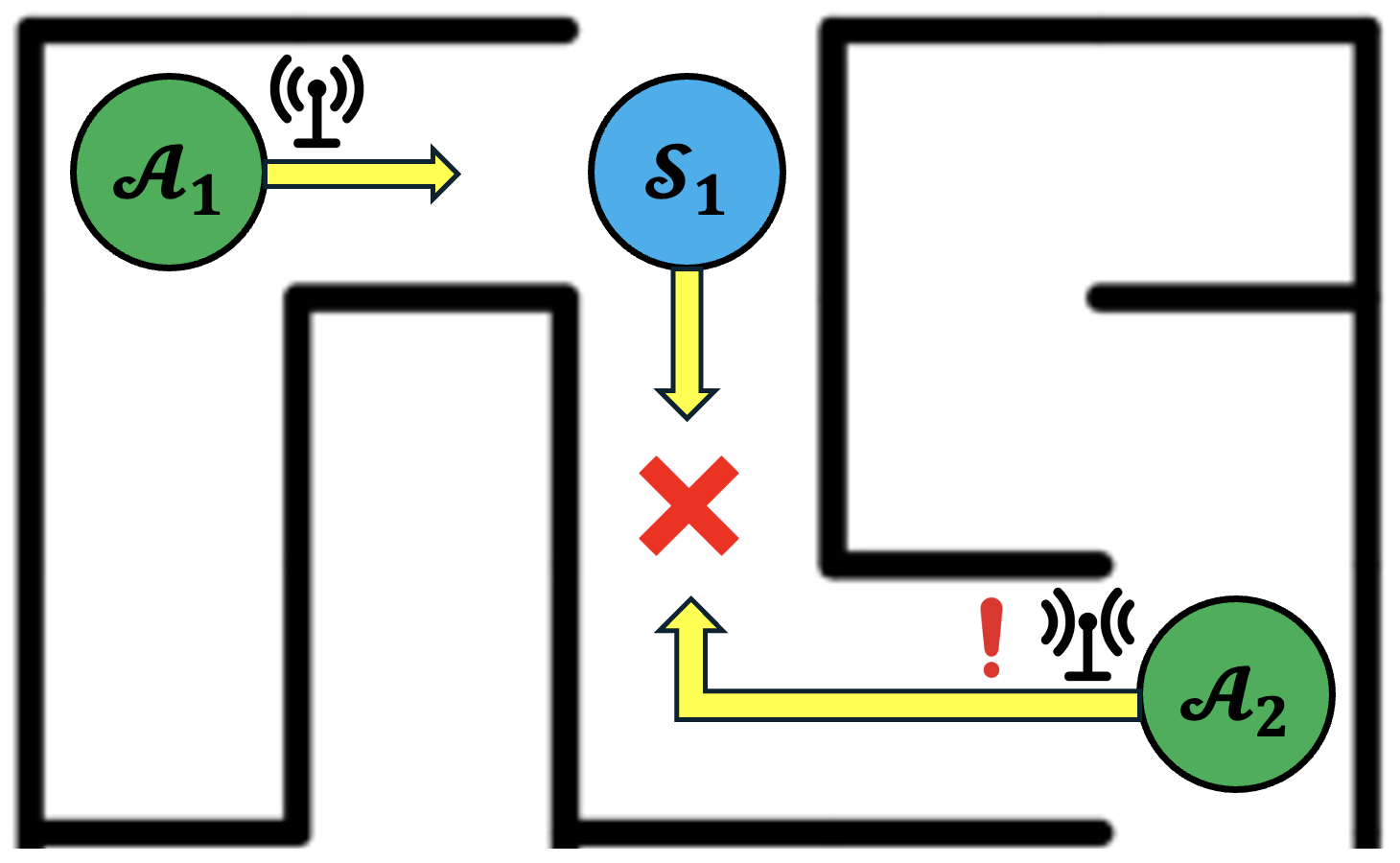}
\caption{As robots $\mathcal{A}_1$ and $\mathcal{A}_2$ navigate through a warehouse environment, $\mathcal{A}_1$ observes moving obstacle $\mathcal{S}_1$ and shares its observation with $\mathcal{A}_2$ over the air. Consequently, $\mathcal{A}_2$ is made aware of a previously invisible potential collision with $\mathcal{S}_1$ and can plan accordingly.\textbf{}}
\end{figure}

There has been significant prior work on decision-theoretic approaches for communication in multi-agent systems. 
Many methods evaluate potential communications based on their expected effect on team reward \cite{carlin1,carlin2}. 
However, computing this value directly for general multi-agent models like Dec-MDPs or Dec-POMDPs is prohibitively complex \cite{pynadath,goldman}. 
The complexity arises from the need to reason about the effects of communication on future team behavior, which requires modeling the beliefs and potential actions of all agents.
Most existing methods attempting to solve this problem focus primarily on deciding the frequency of communication, e.g. via expected value techniques, assuming messages contain an agent's entire observation history \cite{becker,unhelkar}. 
However, this is impractical under real-world bandwidth constraints, where robots must carefully select which observations are most valuable to share.


To address the problem of minimizing the actual content of messages subject to bandwidth constraints, some initial work \cite{roth1,roth2} used greedy optimization to select observations to include in messages.
While efficient, the method suffers poor scalability to more complex problems due to quickly decaying solution quality.
More recent approaches have also explored using deep reinforcement learning to learn communication policies \cite{foerster,sukhbaatar}. 
While capable of scalability to very large scenarios, such learning-based approaches are usually not generalizable to a broader class of multi-robot navigation problems without extensive training.
Thus, prioritizing generalizability over radical scalability, we are motivated to pursue an optimization-based approach for determining communications.

Most recently, Marcotte et al. \cite{marcotte} proposed a method for optimizing multi-robot communication under bandwidth constraints using a bandit-based combinatorial optimization technique to select observations for inclusion in messages. 
Their approach, OCBC, represents the state of the art due to its efficient consideration of observation utility.
However, the formulation relies on a broadcast-based communication scheme where messages are shared with all agents simultaneously. 
In practice, the utility of a particular piece of information for decision-making is often highly subjective to each individual agent.
For example, the observation of some obstacle might only be useful to agents whose planned paths potentially run into that obstacle, and irrelevant to those far away.
In that case, by sending messages only to those who need them, we can save communication bandwidth as well as computation power without sacrificing performance.
Thus, we are motivated to formalize the above intuition into a pairwise communication scheme.

In this paper, we present \textbf{Intelligent Knapsack (iKnap)}, a novel optimal communication scheme for multi-robot observation sharing during navigation in dynamic environments under bandwidth constraints. 
Our method builds on insights from inferential statistics and combinatorial optimization to develop a principled yet tractable solution suitable for real-world implementation.
Our key contributions are:
\begin{itemize}
    \item Modeling pairwise communication in a multi-robot system as graph-based multi-agent inference.
    \item Formulation of the optimal communication problem as a 0/1 knapsack problem, enabling direct and efficient combinatorial optimization of the tradeoff between information value and communication resource usage.
    \item Comprehensive empirical analysis of the proposed method's navigation performance in increasingly complex environments and adaptation to various resource levels and uncertainty conditions.
\end{itemize}

The rest of our paper is organized as follows.
Section \ref{probform} formulates the limited bandwidth observation sharing problem.
Section \ref{method} presents our 0/1 knapsack approach to solving this problem.
Section \ref{results} discusses the results of simulated experiments using our approach.
Finally, Section \ref{conclusion} concludes our study and presents future directions.

\section{Problem Formulation}\label{probform}

We are given a multi-robot system of $n$ agents navigating through an environment where each agent has some goal location it must travel toward.
The environment contains $m$ subjects, i.e. dynamic obstacles, that the agents wish to observe in order to improve decision-making during their navigation task.
The agents and subjects together operate in a state space $\mathcal{X}$, where $\boldsymbol{x}^A_i \in \mathcal{X}$ denotes the state of the $i$-th agent and $\boldsymbol{x}^S_k \in \mathcal{X}$ denotes the state of the $k$-th subject.
Moreover, the $i$-th agent's belief about the $k$-th subject's state variable at time $t$ is modeled by a multivariate probability density function $\boldsymbol{\Phi}^t_{ik} : \mathcal{X} \mapsto \mathbb{R}$.
We shall assume that all agents' states are known by each other.

Each agent $i$ continually obtains observations of the states of the observable subjects in its environment.
The observation $\boldsymbol{\mu}_{ik}^t$ of subject $k$'s state at time $t$ is corrupted with some inherent noise $\boldsymbol{\sigma}_{ik}^{t*}$, and agent $i$ self-assigns some perceived uncertainty $\boldsymbol{\sigma}_{ik}^{t}$ to the observation.
This represents real-world perception systems well, as a robot is obviously incapable of assessing the exact measurement error of its own sensors.
As an agent accumulates more observations about a subject, it updates its beliefs about that subject $\boldsymbol{\Phi}^t_{ik}$ accordingly.

Agents can share observations with each other over the air, i.e. at some time $t$ during navigation, agent $a$ can share the observation $(\boldsymbol{\mu}^t_{ah}, \boldsymbol{\sigma}^t_{ah})$ regarding subject $h$'s state with agent $b$, thus updating $b$'s belief $\boldsymbol{\Phi}^t_{bh}$ accordingly.
At any $t$, the multi-robot system has a total communication bandwidth limit of $B$.
Let communication between $a$ and $b$ regarding subject $h$ at time $t$ occupy some bandwidth $\beta^t_{abh}$.
Moreover, denote the set of all communications $(a,b,h)$ occurring at time $t$ as $\mathcal{C}^t_* := \{(a_1,b_1,h_1),...,(a_{N*},b_{N*},h_{N*})\}$.
Then we have $\sum_{(a,b,h) \in \mathcal{C}^t_*} \beta^t_{abh} \leq B$ for all $t$.

The agents aim to minimize some navigation cost function $J(\boldsymbol{x^A})$ as they travel toward their goals from time $t=0$ to $t=T$.
The agents communicate observations with periodicity $\Delta t$, and this communication strategy potentially improves navigation performance as agents are made more aware of their environment, e.g. collision avoidance.
Thus, the agents must solve the following optimization problem for communication-enabled navigation:
\begin{align}
    &\hspace{-2pt}\min \hspace{5pt} \sum_{t=0}^T J(\boldsymbol{x^A}) \hspace{2pt} \label{eq:prob} \\
    &\text{s.t.} \sum_{(a,b,h) \in \mathcal{C}_*^t} \hspace{-8pt} \beta^t_{abh} \leq B \nonumber \\ &\forall t \in (0, \Delta t,..., T-\Delta t, T) \nonumber
\end{align}
In the following section, we will solve this problem by investigating the construction of $\mathcal{C}^t_*$ more closely.

\section{Intelligent Knapsack Approach}\label{method}
In this section, we present the the iKnap optimal communication scheme in detail.
We first formulate a pairwise communication scheme for the multi-robot system. 
We then present the combinatorial optimization problem for selecting observations to share and detail solving this problem via 0/1 knapsack.

\subsection{Communication Infrastructure}
For more efficient processing of optimal communications, we  establish a centralized communication infrastructure $\mathcal{I}$ to allocate bandwidth in coordinating precise pairwise communications between agents, as opposed to a broadcast-based infrastructure.
 Each agent $i$ then periodically uploads its state $\boldsymbol{x}^A_i$, observations $(\boldsymbol{\mu}_i, \boldsymbol{\sigma}_i)$, and beliefs $\boldsymbol{\Phi}_{i}$ to $\mathcal{I}$, which can then relay this information to the other agents.
We assume that this upload process occupies some fixed bandwidth.
Then the limited remaining bandwidth $B$ on $\mathcal{I}$ is utilized for sharing observations between the agents.
In particular, $B$ is an upper bound on the volume of total simultaneous communication occurring through $\mathcal{I}$ at any given timestep.

\subsection{Pairwise Communication Scheme}
At each timestep $t$, say we have a set of $N$ potential communications $\mathcal{C}^t = \{C^t_1,..., C^t_N\}$ to choose from, and some subset $\mathcal{C}^t_*$ is the set of communications to be actually executed by $\mathcal{I}$ at time $t$.
Each potential communication $C_i := (a_i, b_i, h_i, \beta_i, \theta_i)$ is directed from agent $a_i$ to agent $b_i$, concerning subject $h_i$, and has an associated bandwidth cost $\beta_i$ and associated \textit{utility} $\theta_i$.
One may consider $\theta_i$ as a measure of how ``useful'' $a_i$'s observation is for $b_i$'s decision-making.
While it is difficult to measure utility objectively, various principled heuristics have been introduced in the literature \cite{williamson1,williamson2}.
For potential communication $i$ from agent $a$ to agent $b$ regarding subject $h$, the utility heuristic function is $\theta_i(\boldsymbol{x}^A_b, \boldsymbol{x}^S_h, \boldsymbol{\Phi}^A_a, \boldsymbol{\Phi}^A_b)$.
Note that our method can be generalized to a decentralized formulation to satisfy individual robots' bandwidth constraints, where the utility heuristic function is $\theta_i(\boldsymbol{x}^A_b, \boldsymbol{x}^S_h, \boldsymbol{\Phi}^A_a)$, assuming that state communication between robots occupies a fixed bandwidth.

In practice, $\mathcal{C}$ might contain all pairs of agents $(a_i, b_i)$, or be restricted to only those certain pairs in the system which are able to communicate.
Agent communicability can be represented graphically, hence transitivity may apply, i.e. if $a$ communicates with $b$ with bandwidth $\beta_1$ and $b$ communicates with $c$ with bandwidth $\beta_2$, then $a$ can communicate with $c$ indirectly with bandwidth $\beta_1 + \beta_2$.

One key assumption we make when formulating this communication scheme is that the bandwidth cost of a broadcast can be approximated as the sum of the bandwidth costs of each individual pairwise communications. This assumption is well-supported by empirical studies in multi-robot communications. For example, broadcast transmissions require power proportional to maximum sender-receiver distance, while directed communications optimize based on specific distances \cite{yan}. Moreover, broadcasts suffer increased packet loss in larger teams \cite{amato}, with pairwise schemes achieving significant bandwidth reduction compared to broadcast approaches \cite{queralta}. These findings from real-world deployments validate our modeling approach.

\begin{figure*}[t]
    \centering
    \subfloat[Observation Uploading]{\label{fig:ex_a}\includegraphics[width=.3\linewidth]{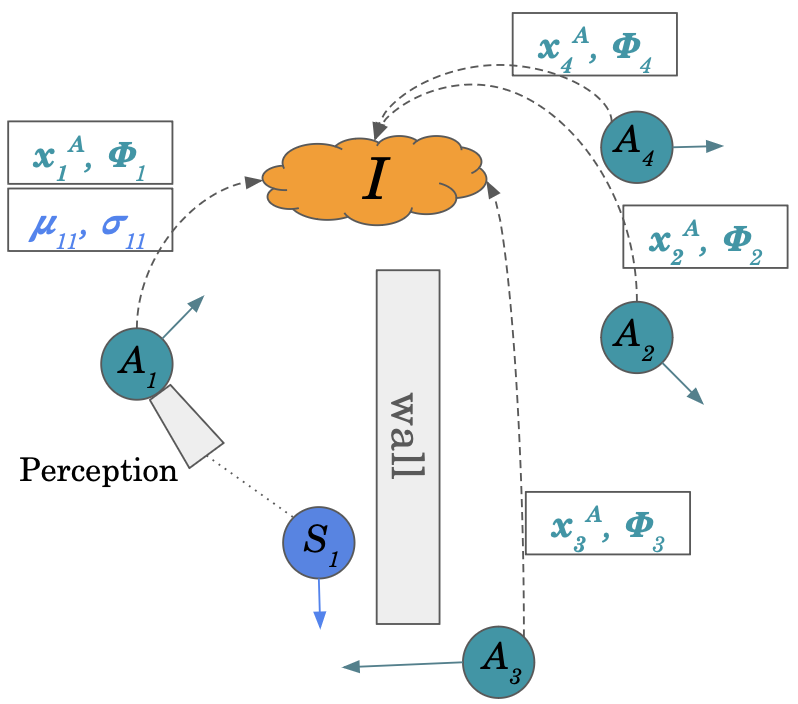}}
    \hspace{1em}
    \subfloat[iKnap Pairwise Communication]{\label{fig:ex_b}\includegraphics[width=.3\linewidth] {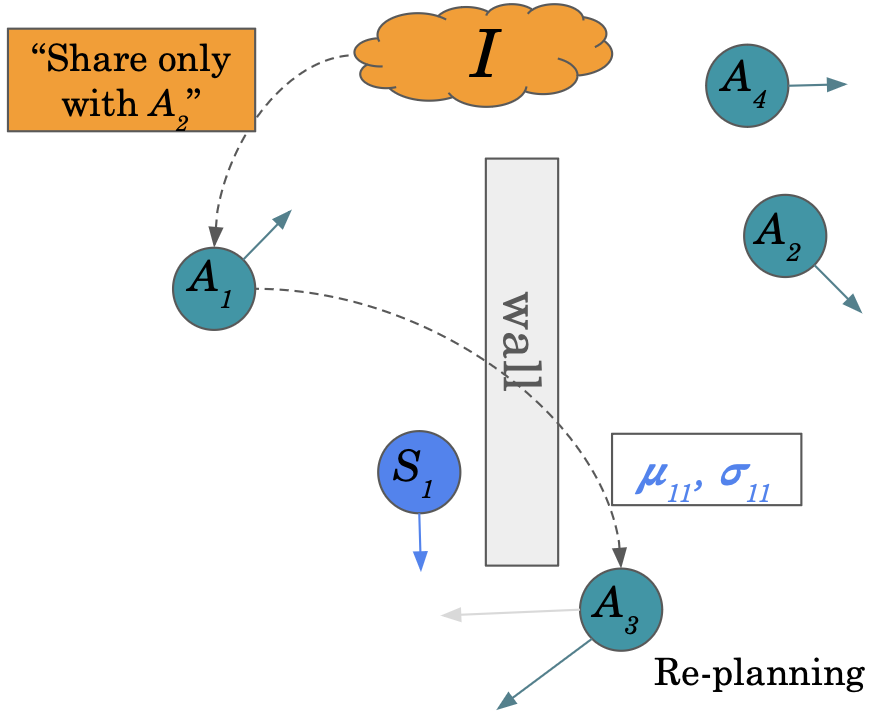}}
    \hspace{1em}
    \subfloat[Broadcast-Based Scheme]
    {\label{fig:ex_c}\includegraphics[width=.3\linewidth]{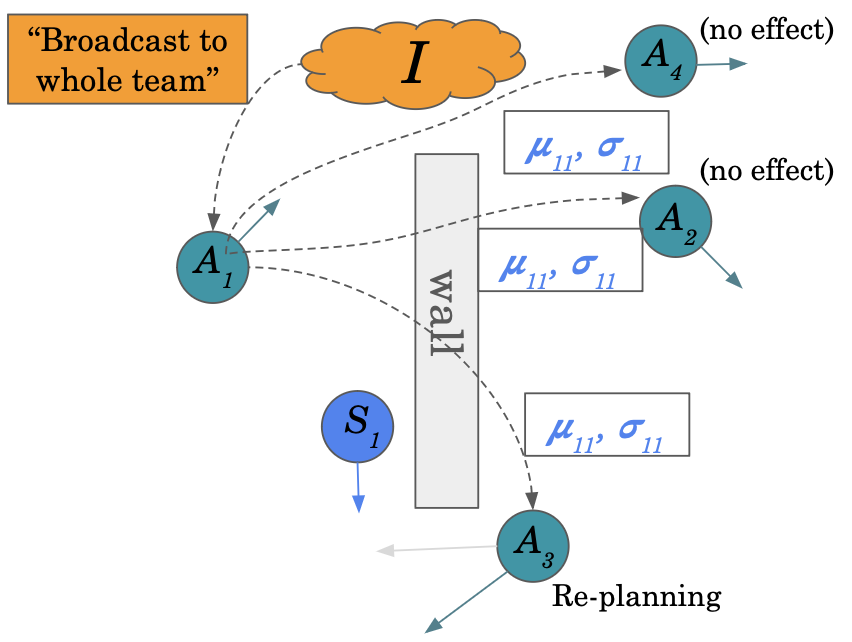}}
    \caption{Example scenario involving a robot team $A_1, A_2, A_3, A_4$ and a dynamic obstacle subject $S_1$ only visible by $A_1$ due to a wall obstruction. There is an impending collision between $A_3$ and $S_1$. In (a), all agents routinely upload their states and beliefs to the infrastructure $\mathcal{I}$, and $A_1$ additionally uploads its observation of $S_1$ to $\mathcal{I}$. Following this process, the proposed pairwise communication scheme is presented in (b), where $A_1$'s observation is shared only with $A_3$ to enable safe re-planning around $S_1$ while maintaining efficient communication. Alternatively, (c) depicts the traditional broadcast-based communication scheme, where $A_1$'s observation is shared with the entire robot team, despite the information's irrelevance to the decision-making of $A_2$ and $A_3$.}
    \label{fig:example}
\end{figure*}

\subsection{Combinatorial Optimization Problem}
It follows that at each timestep $t$, we must select some subset of $\mathcal{C}^t$ to construct $\mathcal{C}^t_*$ for communication across $\mathcal{I}$.
We shall denote this selection as a boolean vector $\boldsymbol{\psi}$, where $\psi_i = 1$ implies $C^t_i \in \mathcal{C}^t_*$ and $\psi_i = 0$ implies $C_i \notin \mathcal{C}^t_*$.
At any given timestep, we seek to find the optimal communication subset $\boldsymbol{\psi}^*$ that maximizes the cumulative utility of the communications across $\mathcal{I}$ while satisfying the bandwidth limit $B$.
Thus, at each timestep we can run the following optimization on $\mathcal{I}$:
\begin{align}
    &\boldsymbol{\boldsymbol{\psi}}^* = \operatorname*{argmax}_{\boldsymbol{\psi}} \hspace{5pt} \boldsymbol{\psi}^T \boldsymbol{\theta}\label{eq:opt} \\
    \text{s.t.} \hspace{15pt} &\boldsymbol{\psi}^T \boldsymbol{\beta} \leq B, \nonumber \\
    &\boldsymbol{\psi} \in \{0, 1\}^N \nonumber
\end{align}
where $\boldsymbol{\theta} := \{\theta_1,...,\theta_N\}$ and $\boldsymbol{\beta} := \{\beta_1,...,\beta_N\}$.

This problem takes the form of a 0/1 knapsack problem, which is solvable in pseudo-polynomial time via dynamic programming.
In particular, the algorithm finds a subset of $N$ items, each with some weight and value, to place in a knapsack of weight capacity $B$ such that the cumulative value is maximized in $O(NB)$ time.
In our case, $N$ is $O(n^2)$, yielding a time complexity of $O(n^2B)$.
For a depiction of the overall observation sharing process, see Algorithm \ref{alg:obs_sharing}.
For an example of the observation uploading process and a comparison of pairwise and broadcast-based communication schemes, see Figure \ref{fig:example}.


\section{Experimental Results}\label{results}
\begin{figure}
    \centering
    \includegraphics[width=\linewidth]{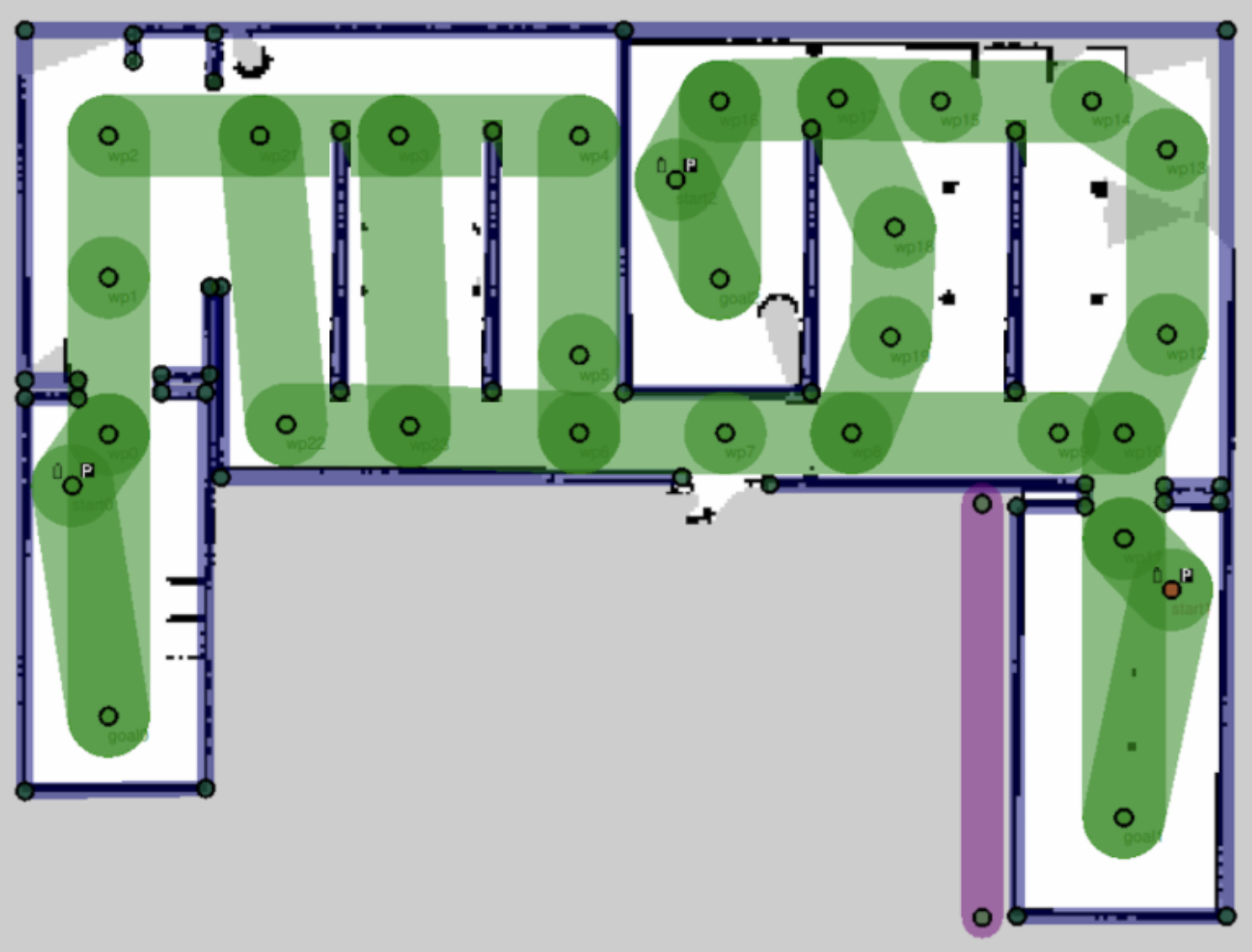}
    \caption{Example warehouse environment generated with the Open-RMF Traffic Editor tool.}
    \label{fig:map}
\end{figure}

\subsection{Simulation Setup}
We design a high-fidelity multi-robot warehouse simulation environment using the Robot Operating System (ROS 2) \cite{ros2} and the Open Robotics Middleware Framework (RMF) \cite{rmf}. 
We choose RMF for its comprehensive capabilities in simulating the full multi-robot navigation stack.
The environment consists of a warehouse with $n$ robots, $m$ dynamic obstacles (simulated human workers), and various static obstacles (walls, shelves) to navigate around.
The simulation leverages RMF's \texttt{traffic\_editor} package to generate randomized warehouse environments with realistic layouts. Robots follow paths computed by a grid-based A* planner and use the RMF \texttt{slotcar} plugin for dynamics and control. This plugin provides a lightweight but realistic model of differential-drive robot kinematics.
Human workers are simulated as dynamic obstacles using RMF's \texttt{CrowdSim} package, which provides realistic human movement patterns in warehouse environments. The humans follow randomized paths between different locations in the warehouse, creating dynamic navigation challenges for the robots.
All simulations are run using $\Delta t = 0.01$ s.
For an example warehouse environment map generated by RMF, see Figure $\ref{fig:map}$.

\begin{algorithm}[t]
\caption{Observation sharing}
\label{alg:obs_sharing}
\begin{algorithmic}[1]
\State \textbf{Input:} Set of agents $\mathcal{A} = \{a_1, a_2, ..., a_n\}$, set of subjects $\mathcal{S} = \{s_1, s_2, ..., s_m\}$, bandwidth limit $B$, communication period $\delta t$
\State \textbf{Output:} Updated beliefs $\boldsymbol{\Phi}$ for each agent

\For{each time step $t = 0, \delta t, 2\delta t, ..., T$}
    \For{each agent $i \in \mathcal{A}$}
        \State $\boldsymbol{\mu}_{ik}^t  \gets \textsc{ObtainObservations}(i, \mathcal{S}, t)$
        \State $\boldsymbol{\Phi}_{ik}^t \gets \textsc{UpdateBelief}(i, \boldsymbol{\mu}_{ik}^t)$
    \EndFor

    \State $\mathcal{C}^t \gets \{\}$  

    \For{each pair of agents $(a, b) \in \mathcal{A} \times \mathcal{A}, a \neq b$}
        \For{each subject $h \in \mathcal{S}$}
            \State $\beta_{abh} \gets \textsc{BandwidthCost}(a, b, h)$
            \State $\theta_{abh} \gets \textsc{Utility}(a, b, h, \boldsymbol{x}^A_b, \boldsymbol{x}^S_h, \boldsymbol{\Phi}^A_a, \boldsymbol{\Phi}^A_b)$
            \State \textsc{Push}$(\mathcal{C}^t, (a, b, h, \beta_{abh}, \theta_{abh}))$
        \EndFor
    \EndFor

    \State $\mathcal{C}^t_* \gets \textsc{SolveKnapsackProblem}(\mathcal{C}^t, B)$

    \For{each selected communication $C_{abh}^* \in \mathcal{C}^t_*$}
        \State \textsc{ShareObservation}($a, b, h$)
        \State $\boldsymbol{\Phi}_{bh}^t \gets \textsc{UpdateBelief}(b, \boldsymbol{\mu}_{ah}^t)$
    \EndFor
\EndFor
\end{algorithmic}
\end{algorithm}

\subsection{Perception Design}
At each timestep, each robot obtains a noisy observation of each \textit{visible} dynamic obstacle's state (position and velocity). An obstacle is considered visible to a robot if no walls or shelves exist between them, accurately modeling real-world occlusion scenarios. The observation noise is modeled as a Gaussian distribution with standard deviation
\begin{equation}
    \boldsymbol{\sigma}_{ik} = \frac{\alpha}{||\boldsymbol{x}^A_i - \boldsymbol{x}^S_k||^2},
\end{equation}
for tuning parameter $\alpha$, reflecting the realistic property that observations become less accurate with increasing distance.

Each robot maintains a belief state about each dynamic obstacle using Bayesian inference with Gaussian prior and likelihood, assuming constant velocity dynamics. When a robot receives a new observation (either from its own sensors or shared by another robot), it updates its belief accordingly.

\subsection{Communication Infrastructure}
We implement a centralized communication infrastructure that collects observations from all robots, determines optimal sharing strategies, and distributes selected observations. The infrastructure models realistic communication constraints through several key factors. The bandwidth required for communication between robots $a$ and $b$ is modeled as a function of their Euclidean distance, occlusions, and message size:
\begin{equation}
    \beta_{ab} = \beta_{base} + \beta_{dist} \cdot ||x_a - x_b|| + \beta_{occl} \cdot \mathbb{I}_{occl}(a,b)
\end{equation}
where $\beta_{base}$ is a base cost, $\beta_{dist}$ is a distance-dependent factor, and $\beta_{occl}$ is an additional cost applied when occlusions exist between robots. The system also models communication latency as a configurable delay between when a message is sent and when it is received, simulating real-world network conditions. Additionally, a global constraint on the total bandwidth available per communication cycle is implemented to reflect realistic limitations.

\subsection{Utility Heuristic Design}
The utility heuristic provides an estimate of how useful an observation from robot $a$ regarding subject $h$ would be to robot $b$. We consider two primary factors in this calculation. First, we measure the improvement in $b$'s belief about $h$ that would result from $a$'s observation, using the Kullback–Leibler divergence between prior and posterior distributions of $\boldsymbol{\Phi}_{bh}$, denoted as $\kappa_{abh}$. Second, we assess the relevance of knowledge about $h$'s state to $b$'s navigation task by forward-simulating trajectories and computing the inverse-squared minimum distance between $b$ and $h$ across a prediction horizon of $t_H$ seconds, denoted as $\tau_{abh}$. The utility for potential communication $C_i$ associated with $(a,b,h)$ is given by
\begin{equation}
    \theta_i = p_1 \kappa_{abh} + p_2 \tau_{abh}
\end{equation}
for tuning parameters $p_1$ and $p_2$ that normalize and determine the relative importance of each metric. 
Intuitively, $\kappa_{abh}$ measures the improvement in $b$'s knowledge, and $\tau_{abh}$ measures the applicability of $b$'s knowledge.
Taken together, these two factors form a reasonable heuristic for the true ``utility'' of a given communication.
We note that there are many other ways to choose the utility heuristic, e.g. via learning-based approaches.

\subsection{Planner and Controller Design}\label{pcd}
Each robot plans a path from its start location to its goal location using a grid-based A* algorithm that accounts for static obstacles. The robots follow these paths using the RMF \texttt{slotcar} plugin, which implements an adapted model of differential-drive robot dynamics.
For collision avoidance with dynamic obstacles, each robot leverages its belief about nearby obstacles to guide naive trajectory prediction with a constant-velocity assumption. If a collision is predicted, the robot decelerates linearly or stops until its confidence threshold path is clear is satisfied. The trajectory predictor is implemented with a variable prediction horizon parameter.
This implementation rewards robots having greater confidence in their observations, as this directly improves navigation outcomes by reducing unnecessary slowdowns.

\subsection{Dynamic Obstacles Design}
To simulate human workers in the warehouse environment, we leverage RMF's \texttt{CrowdSim} package which integrates the Menge library for realistic human movement. Dynamic obstacles are modeled as circular agents with configurable radii (0.25m) navigating through a graph-based global plath planner and Optimal Reciprocal Collision Avoidance (ORCA) \cite{orca} for local collision avoidance. Human movements follow a Finite State Machine where agents transition between goal states based on proximity or time conditions. Each human has individualized parameters including maximum acceleration (5 m/s$^2$), maximum angular velocity (360 deg/s), preferred speed (1.5 m/s), and maximum speed (2 m/s). This implementation creates non-deterministic human movements that respond dynamically to both the environment and other agents, providing realistic warehouse interaction scenarios that challenge the robots' navigation capabilities.

\subsection{Baselines and Evaluation Metrics}
We compare the iKnap framework to four baseline approaches. The No Communication baseline has robots rely solely on their own observations without sharing information. The Random Broadcast approach has robots randomly decide whether to broadcast their observations to all other robots with a fixed probability. In the Random Pairwise method, robots randomly decide whether to share observations with each specific robot with a fixed probability. Finally, OCBC is a state-of-the-art observation sharing method proposed in \cite{marcotte} that uses a broadcast-based communication scheme with a bandit-based combinatorial optimization approach for utility-based observation selection.



We measure algorithm performance via two primary metrics: makespan, which is the time until the last robot reaches its goal and directly reflects the overall efficiency of the multi-robot system; and optimizer runtime, which is the computational time required by the communication selection algorithm and is important for real-time applications.

All experiments were run on a Linux desktop computer with an Intel Core i7-12700KF CPU and 16GB DDR5 RAM. In the following sections, we first evaluate the capability of iKnap compared to all baselines, and then perform a robustness analysis comparing primarily to OCBC.

\begin{figure*}[t]
    \centering
    \subfloat[Robot/obstacle count $n+m$]{\includegraphics[width=.25\linewidth]{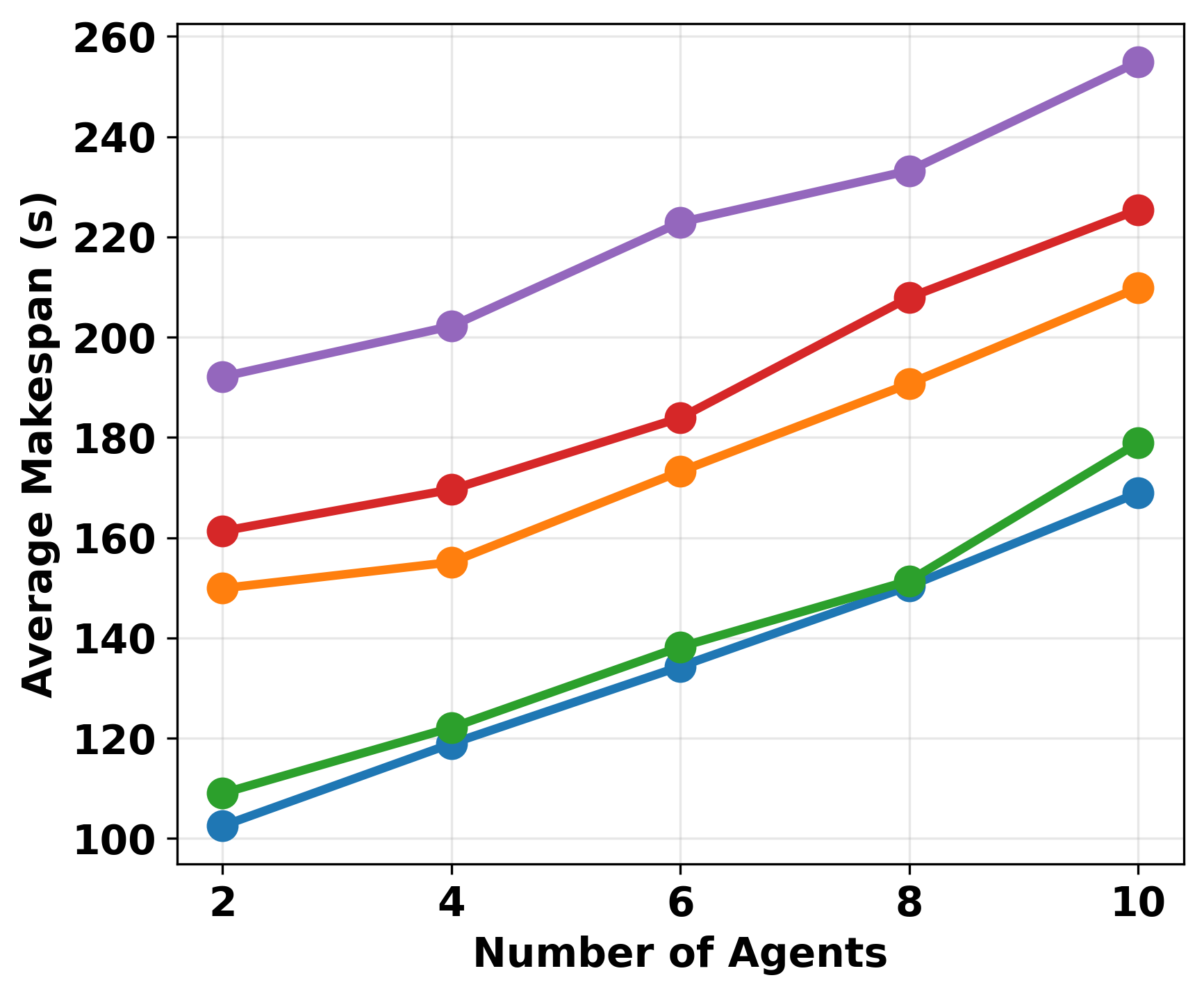}}
    \hfill
    \subfloat[Noise parameter $\alpha$]{\includegraphics[width=.25\linewidth]{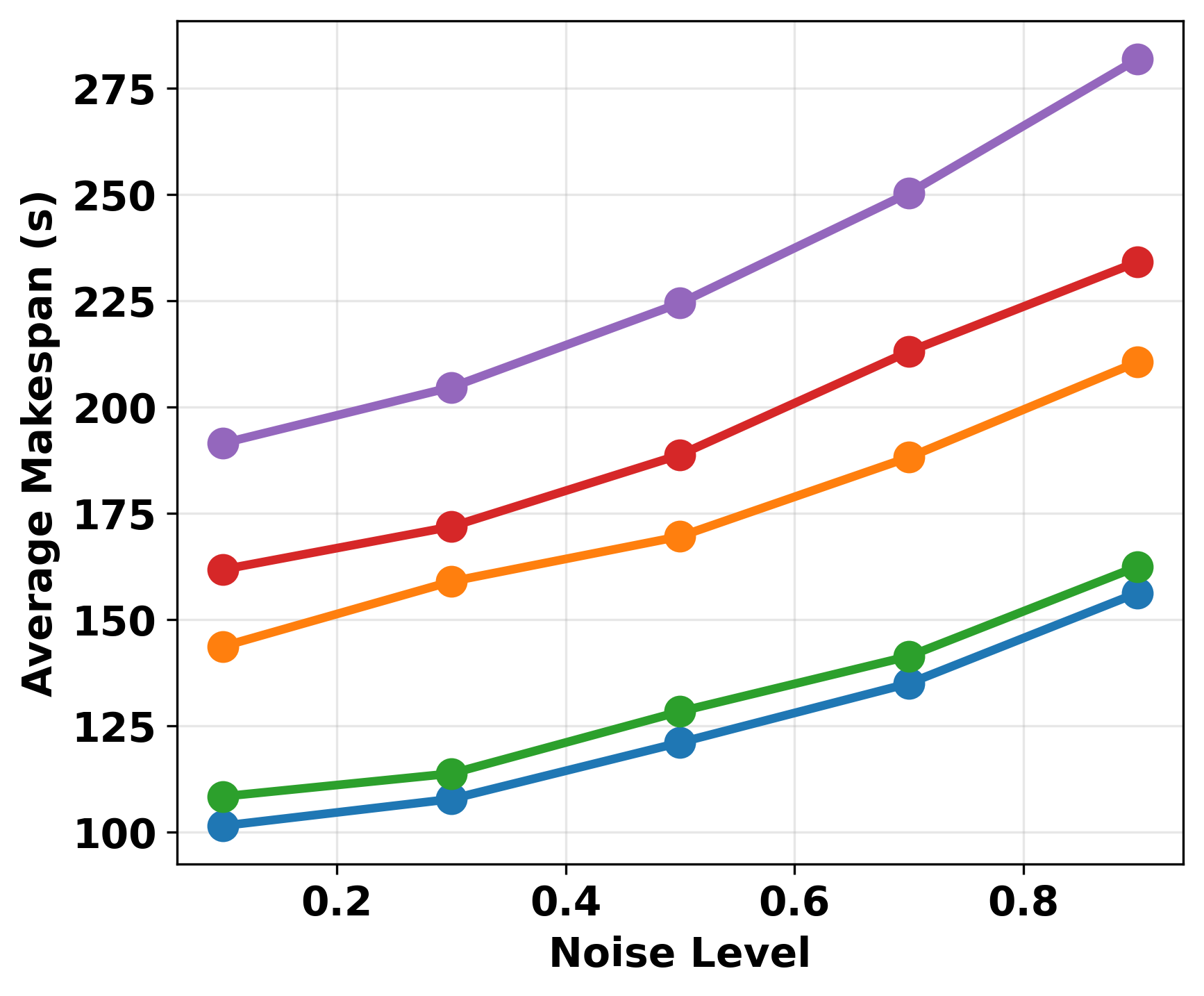}}
    \hfill
    \subfloat[Communication latency]{\includegraphics[width=.25\linewidth]{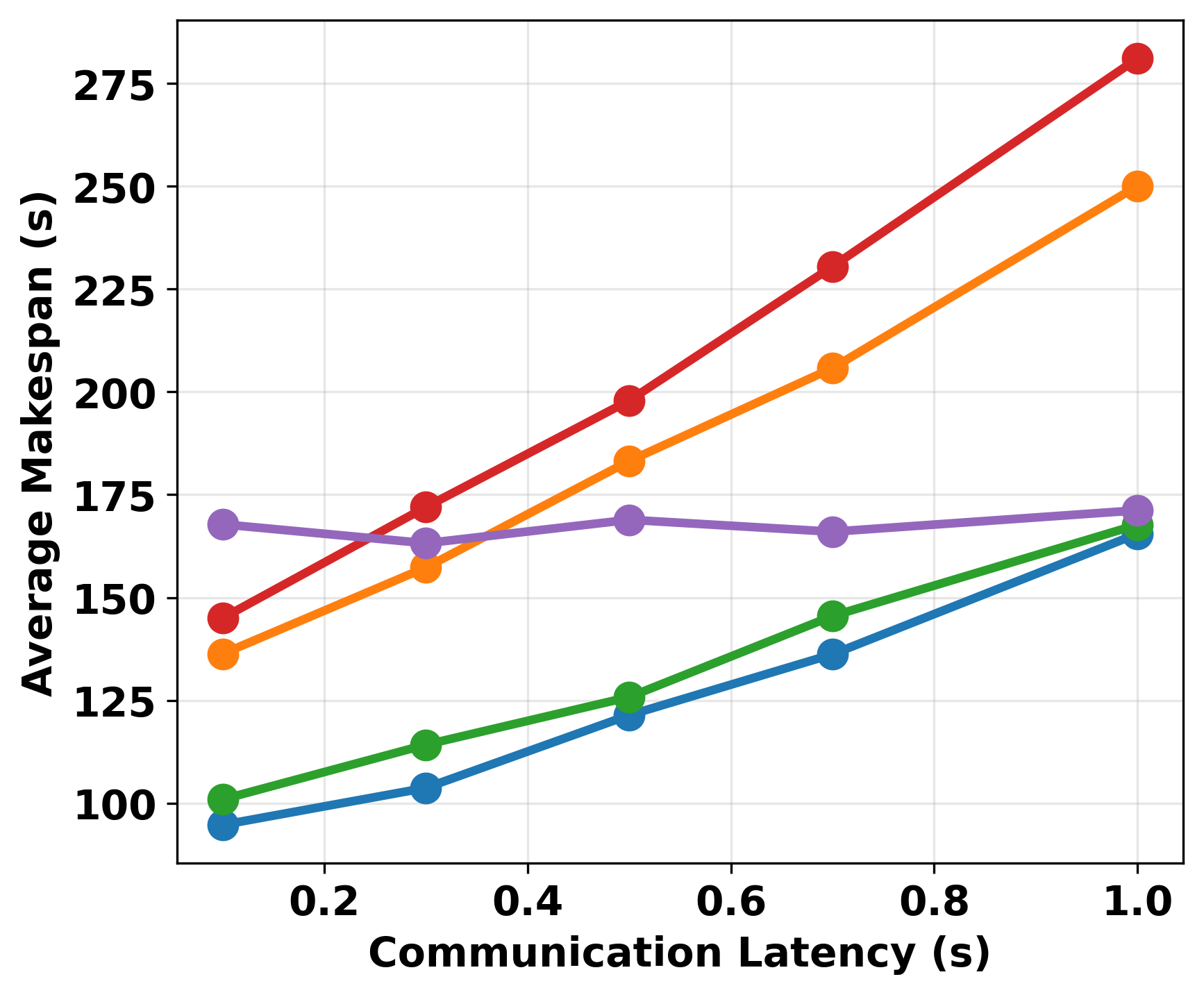}}
    \hfill
    \subfloat[Wall count $w$]{\includegraphics[width=.25\linewidth]{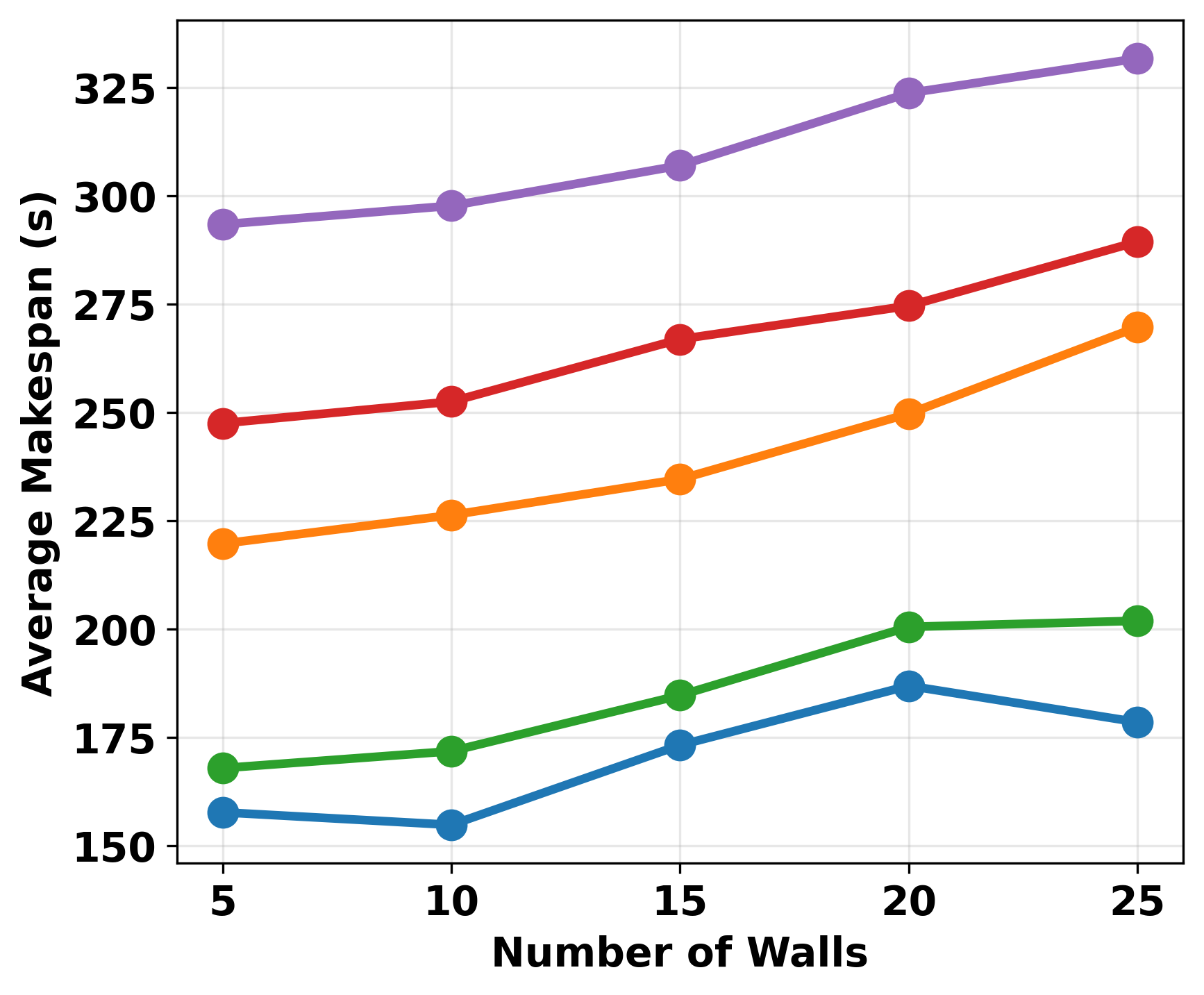}}

    \caption{Capability studies comparing makespan (s) for iKnap (blue), OCBC (green), random pairwise (red), random broadcast (orange), and no communication (purple) baselines. We measure performance vs. (a) robot/obstacle count $n+m$, (b) sensor noise level, (c) communication latency, and (d) environment complexity (number of walls). Each configuration's measurements are averaged across multiple randomized trials.}
    \label{fig:capability}
\end{figure*}

\subsection{Capability Studies}
To assess the capability of iKnap, we measure navigation performance with respect to the size of the communication problem, environment complexity, observation noise, and communication latency. We compare our approach to all baselines across these dimensions.
For each parameter configuration, we average performance across $20$ randomly generated warehouse environment configurations. When altering one parameter, we hold all other parameters constant at half of their maximum tested value.

Figure \ref{fig:capability} displays the results of the capability experiments. These results demonstrate that iKnap consistently outperforms all baselines across all tested conditions. As expected, the no-communication baseline performs worst, while the random communication approaches show modest improvements. OCBC performs significantly better than the random approaches but is consistently outperformed by iKnap.
Notably, randomized pairwise communication poses a consistent advantage over randomized broadcast-based communication, highlighting the fundamental differences between these formulations.

Our observations specifically indicate that our approach offers significant advantages in several key areas. In terms of scalability (Fig. \ref{fig:capability}a), as the number of agents and obstacles increases, all methods show increased makespan due to greater congestion and coordination challenges. However, iKnap scales more gracefully, maintaining a \~10\% advantage over OCBC even at higher agent counts. Regarding robustness to noise (Fig. \ref{fig:capability}b), with increasing sensor noise, all methods degrade in performance. Our approach shows greater robustness, with the performance gap between iKnap and OCBC widening as noise increases, demonstrating the value of selective information sharing under uncertainty. For communication latency (Fig. \ref{fig:capability}c), higher latency negatively impacts all communication-based approaches. Our approach maintains better performance across all latency levels, suggesting more effective prioritization of time-sensitive information. In terms of environmental complexity (Fig. \ref{fig:capability}d), as the number of walls increases, creating more complex navigation scenarios with occlusions, iKnap maintains its advantage, showing \~15\% lower makespan compared to OCBC.

\begin{figure*}[t]
    \centering
    \begin{minipage}{0.25\textwidth}
        \centering
        \includegraphics[width=\textwidth]{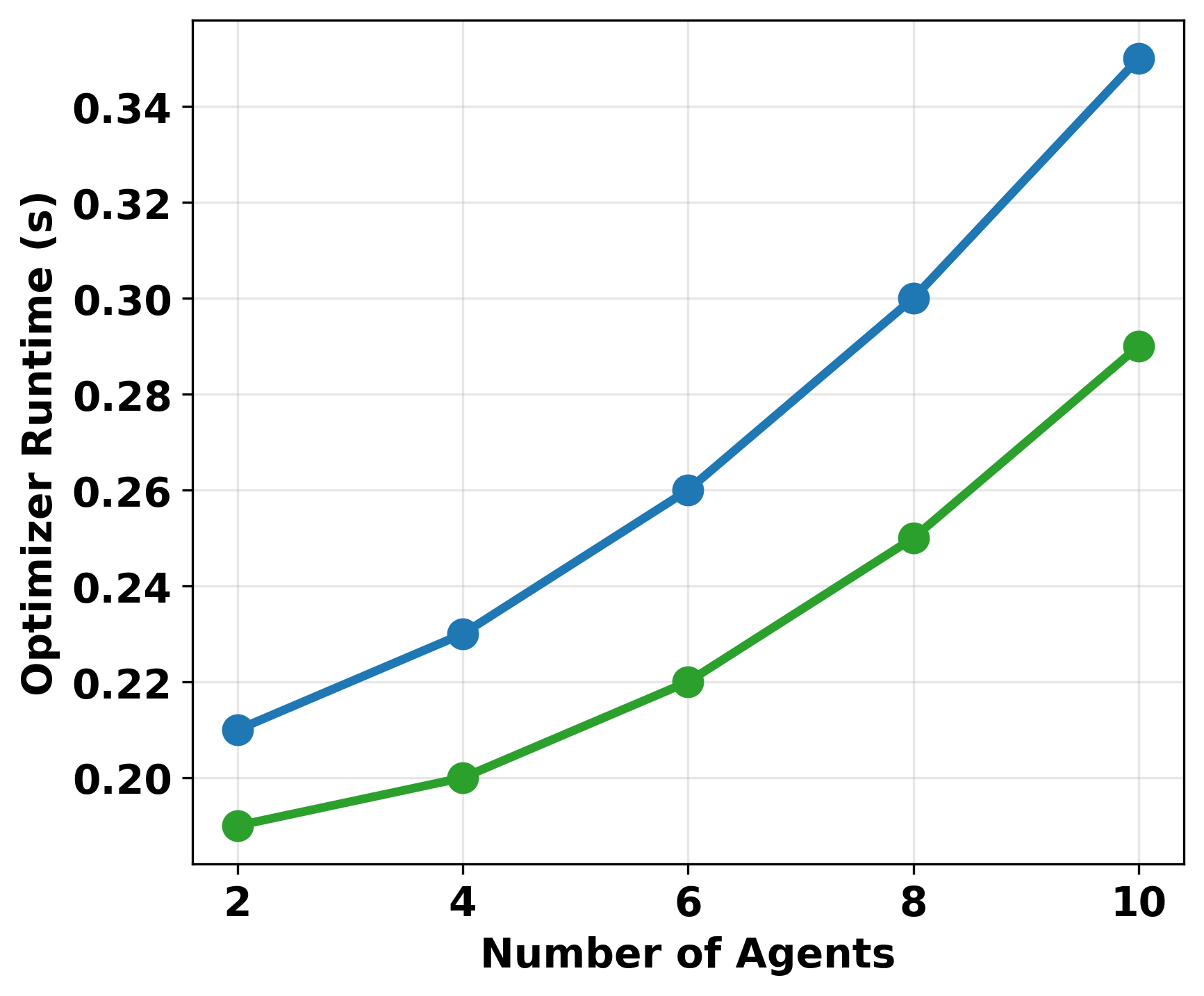}
        \caption{Optimizer runtime (s) vs. agent/obstacle count for iKnap (blue) and OCBC (green).}
        \label{fig:runtime}
    \end{minipage}%
    \hfill
    \begin{minipage}{0.73\textwidth}
        \centering
        \subfloat[Communication frequency (hz)]{\includegraphics[width=0.32\textwidth]{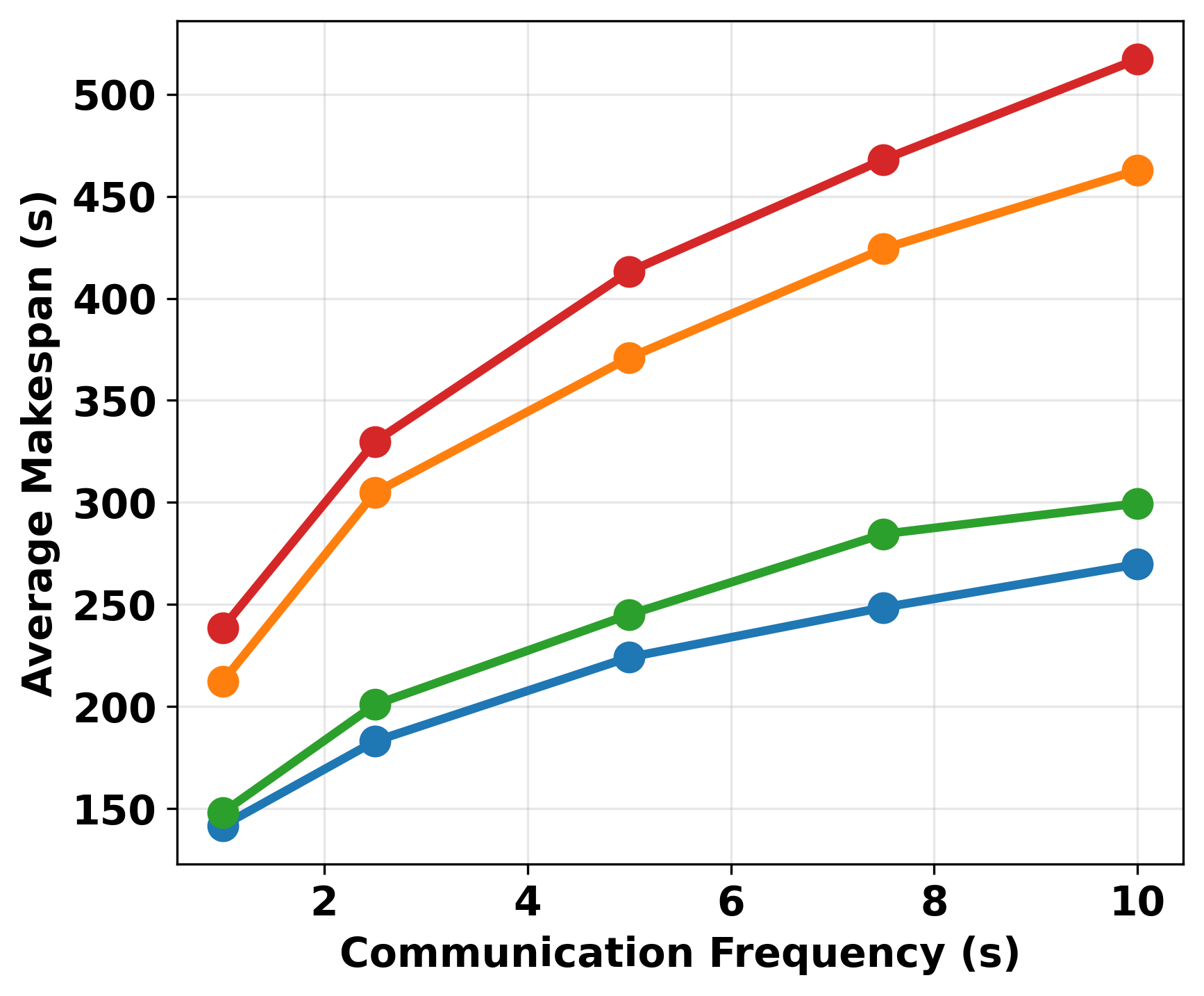}}
        \hfill
        \subfloat[Bandwidth limit]{\includegraphics[width=0.32\textwidth]{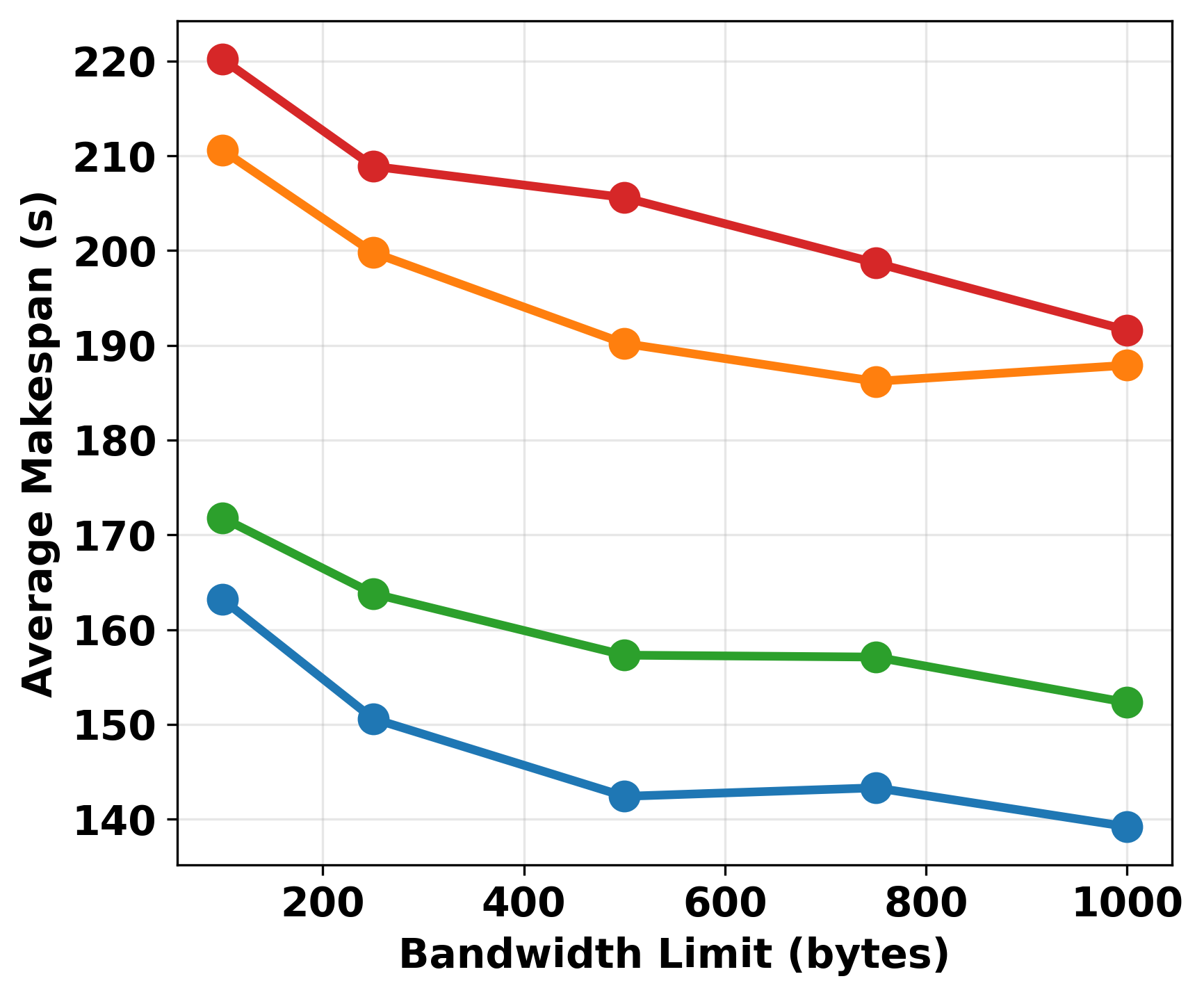}}
        \hfill
        \subfloat[Prediction horizon (s)]{\includegraphics[width=0.32\textwidth]{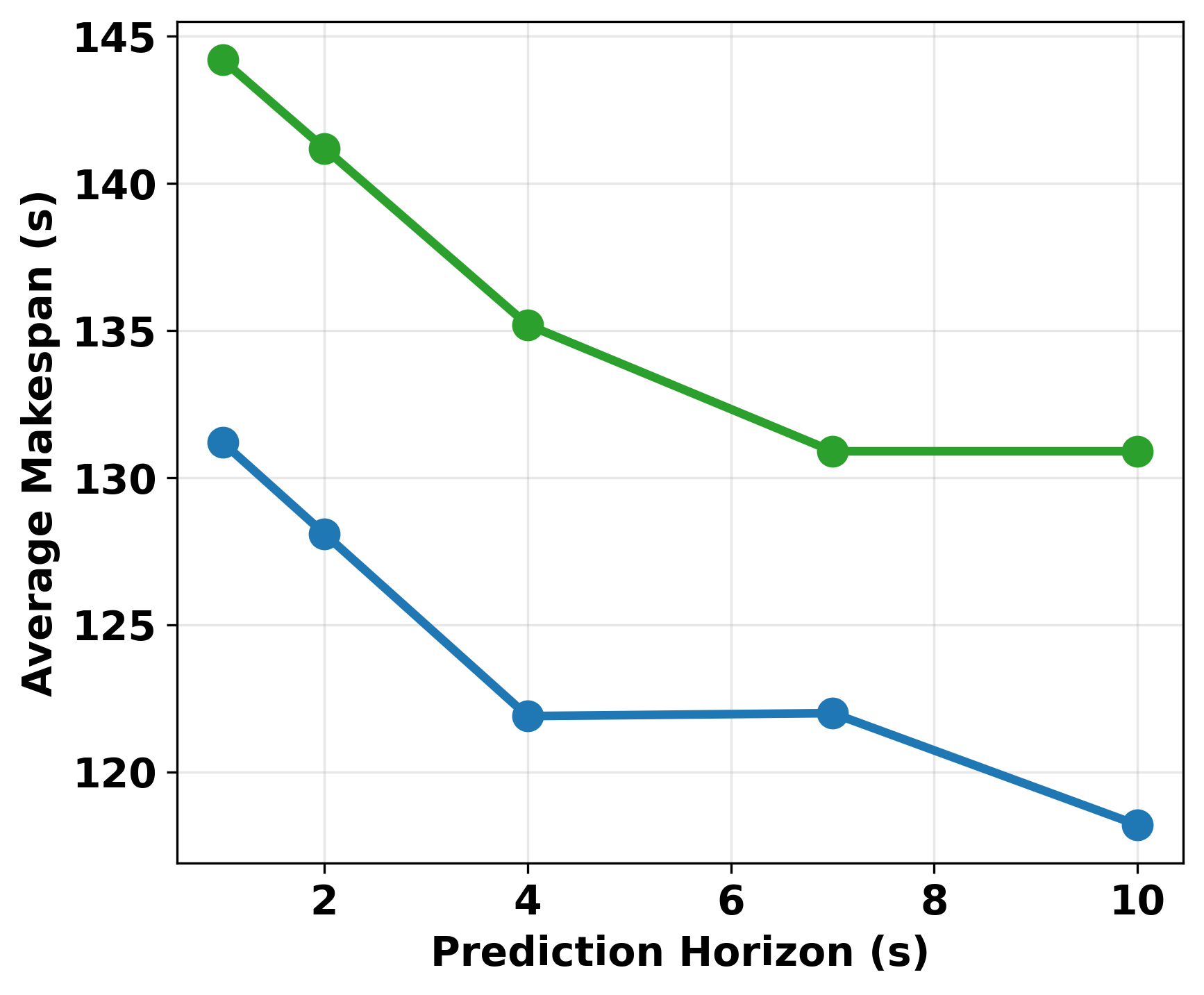}}
        \caption{Robustness analysis comparing makespan (s) for iKnap (blue), OCBC (green), random pairwise (red), and random broadcast (purple) approaches. We measure performance vs. (a) communication frequency, (b) bandwidth limit, and (c) trajectory prediction horizon. Each configuration's measurements are averaged across multiple randomized trials.}
        \label{fig:ablation}
    \end{minipage}
\end{figure*}

Figure \ref{fig:runtime} shows the optimizer runtime for both iKnap and OCBC. Despite the theoretical complexity of the knapsack problem, our implementation exhibits runtime comparable to OCBC, with only a modest increase as the number of agents grows. Even in scenarios with 10 robots and 10 obstacles, the runtime remains well under 0.4 seconds, making it suitable for real-time applications.

\subsection{Robustness Analysis}\label{ablation}
To analyze the robustness of iKnap, we measure navigation performance with respect to communication frequency, bandwidth allowance, and utility heuristic parameters. We compare primarily to OCBC, with the random approaches included for context.

Figure \ref{fig:ablation} displays the results of the robustness experiments. These results show several significant trends in system performance. Regarding communication frequency (Fig. \ref{fig:ablation}a), higher frequency improves performance for all approaches by allowing more timely information sharing. Our approach shows greater improvement with increased frequency, suggesting iKnap selects more valuable information at each communication cycle. For bandwidth limits (Fig. \ref{fig:ablation}b), as the limit increases, all approaches improve, but with diminishing returns. Our approach shows superior performance across all bandwidth limits, with the advantage being most pronounced at lower limits where selective sharing is most critical. In terms of prediction horizon (Fig. \ref{fig:ablation}c), the horizon affects how far ahead robots consider potential interactions with dynamic obstacles. Both iKnap and OCBC show an optimal performance at a medium horizon (around 7 s), with performance degrading at very short horizons annd flattening out at very long horizons. This suggests that very short horizons provide insufficient lookahead, while very long horizons introduce too much uncertainty in predictions.

These robustness studies demonstrate that iKnap utilizes communication resources more effectively than OCBC and the random baselines. The advantage is particularly significant in resource-constrained scenarios (low communication frequency or bandwidth), highlighting the value of the pairwise communication paradigm and knapsack optimization for real-world multi-robot systems where communication resources are often limited.

\section{Conclusion}\label{conclusion}
In this paper, we introduced Intelligent Knapsack (iKnap), an optimal communication scheme designed for multi-robot observation sharing in dynamic environments with bandwidth constraints. 
By modeling communication as graph-based multi-agent inference and framing the optimization problem as a 0/1 knapsack problem, iKnap efficiently balances the tradeoff between information value and communication cost. 
Our empirical analysis shows that iKnap significantly enhances navigation performance, particularly in complex scenarios, while maintaining comparable runtime to existing broadcast-based schemes. 
It also optimizes bandwidth and observational resource usage more effectively, especially in low-resource and high-uncertainty settings. 
These results suggest that iKnap facilitates robust collaboration among multi-robot teams in real-world navigation challenges.
Moreover, when the subjects are human subjects, the proposed method has the potential to improve safety and efficiency in multi-robot multi-human interaction.
Future research will focus on expanding iKnap's compatibility with diverse agent behavior models and integrating learning-based approaches to construct generic communication utility functions for various scenarios.
We also intend to perform real-world tests in factory environments.

\end{document}